\documentclass{article} 
\usepackage{iclr2027_conference,times}

\usepackage{amsmath,amsfonts,bm}

\def\eqref#1{equation~\ref{#1}}

\def\1{\bm{1}}

\DeclareMathAlphabet{\mathsfit}{\encodingdefault}{\sfdefault}{m}{sl}
\SetMathAlphabet{\mathsfit}{bold}{\encodingdefault}{\sfdefault}{bx}{n}

\usepackage{graphicx}
\usepackage{booktabs}
\usepackage{multirow}
\usepackage{array}
\usepackage{makecell}
\usepackage{amssymb} 
\usepackage[table]{xcolor} 
\usepackage{capt-of}
\definecolor{MethodGroupBlue}{RGB}{236,244,250}
\definecolor{MethodSubgroupGray}{RGB}{248,248,248}

\usepackage{hyperref}
\usepackage{url}

\title{WorldFact-Bench: Beyond Image-Internal Plausibility to Image-World Consistency}

\author{
\textbf{Zhuohong Chen}\textsuperscript{1,2,*}
\quad
\textbf{Zhengxian Wu}\textsuperscript{1,2,*}
\quad
\textbf{Yunyao Yu}\textsuperscript{2,*}
\quad
\textbf{Hangrui Xu}\textsuperscript{2}
\quad
\textbf{Zijian Yu}\textsuperscript{1}
\\[0.5em]
\textbf{Hao Tan}\textsuperscript{1}
\quad
\textbf{Zhifang Liu}\textsuperscript{2}
\quad
\textbf{Peng Jiao}\textsuperscript{2}
\quad
\textbf{Jun Lan}\textsuperscript{1,$\dagger$}
\quad
\textbf{Haoqian Wang}\textsuperscript{2,$\dagger$}
\\[0.7em]
\normalfont
\textsuperscript{1}Ant Group
\qquad
\textsuperscript{2}Tsinghua University
}
\iclrfinalcopy
\begin{document}

\maketitle

\fancyhead{}
\renewcommand{\headrulewidth}{0pt}

\begingroup
\renewcommand{\thefootnote}{}
\footnotetext{
\textsuperscript{*} Equal contribution.
\qquad
\textsuperscript{$\dagger$} Corresponding authors.
}
\endgroup

\begin{figure}[!h]
  \centering
  \includegraphics[width=\linewidth]{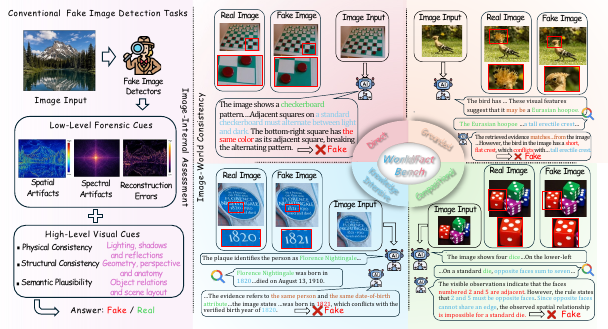}
  \caption{
Comparison between image-internal assessment and image-world consistency verification.
Left: image forensics examines low-level artifacts and high-level
physical, structural, and semantic consistency.
Right: examples from the four verification regimes in WorldFact-Bench
illustrate how visual observations are checked against real-world
facts and rules to identify factual conflicts.
}
  \label{fig:teaser}
\end{figure}

\begin{abstract}
Advances in image generation have made visual authenticity
increasingly difficult to assess.
Although image forensics now examines both generation artifacts
and higher-level visual inconsistencies, a plausible image
can still contradict real-world facts or rules.
We introduce WorldFact-Bench to evaluate \emph{image-world consistency}
from a single image, without a predefined claim or verification target.
The benchmark contains 1,274 source-aligned real--fake pairs
across four verification regimes and ten semantic domains.
Each pair introduces a specific, evidence-supported factual conflict
while seeking to preserve non-target content and visual plausibility.
Images are evaluated independently, and pair accuracy requires
both members of a pair to be classified correctly.
We further propose PERSIST-Agent, which organizes iterative
verification around a persistent state linking candidate facts,
visual observations, evidence, and verification statuses.
This state guides subsequent inspection and retrieval
while retaining unresolved candidates.
With backbone weights fixed, harness self-optimization refines
the agent's prompts and execution rules through validation feedback.
Experiments reveal strong label biases in several detectors
and uneven gains from retrieval.
On the evaluated 8B backbones, PERSIST-Agent improves pair accuracy
over both direct judgment and retrieval-augmented baselines,
while ablations support the role of persistent verification state.
These findings highlight the value of state-guided verification
and the remaining gap between visual plausibility
and factual correctness.
\end{abstract}

\begin{figure*}[t]
    \centering
    \includegraphics[width=\textwidth]{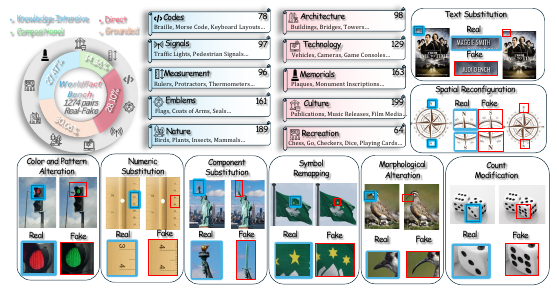}
    \caption{
Overview of WorldFact-Bench.
The dataset contains 1,274 real--fake pairs across four verification
regimes and ten semantic domains.
Left: regime proportions.
Middle: domain counts and representative categories.
Right and bottom: examples of the eight manipulation types,
with blue and red boxes highlighting corresponding factual regions
in the real and fake images, respectively.
}
    \label{fig:dataset-overview}
\end{figure*}

\section{Introduction}
\label{sec:introduction}

Advances in image generation have made synthetic images increasingly
realistic in texture, lighting, and structure.
This progress makes reliable image forensics more challenging.
Existing approaches go beyond low-level generation traces to examine
semantic, structural, and physical cues~\citep{Yan2024ASC}.
Recent multimodal models also reason about visual anomalies
and explain their judgments~\citep{Wen2025SpotTF,Tan2025VeritasGD}.
Yet the absence of obvious anomalies does not establish factual
correctness.
Verification must also consider whether the depicted content
agrees with real-world facts and rules.

For example, a film poster may have a natural layout, clear text,
and convincing portraits, yet credit an actor who did not appear
in the film.
Detecting this error requires checking the film's cast,
not just the appearance of the poster.
We study \emph{image-world consistency}: whether the attributes
and relations depicted in an image agree with the relevant
facts or rules.
Our task asks a model to discover and verify factual conflicts
from a single image, without a predefined claim or verification target.

The challenge is not only to obtain knowledge, but also to decide
what to check and how that knowledge applies to the image.
Factual errors need not produce visible anomalies, so the model
must identify facts worth checking among many plausible details.
Relevant search results alone are not enough.
The model must establish that the evidence concerns the same
entity, attribute, and context before comparing it with
the visual observation.
When several candidates require verification, it must also
keep track of what has been checked, what remains uncertain,
and which observation or search would resolve that uncertainty.
The task therefore requires coordinated visual inspection,
evidence assessment, and verification across multiple steps.

To evaluate this ability, we introduce \textbf{WorldFact-Bench},
with 1,274 source-aligned real--fake image pairs across four
verification regimes and ten semantic domains.
Each pair starts from a source image with a visible fact
supported by traceable evidence.
We introduce a specific factual conflict while seeking to preserve
the remaining content and visual plausibility.
This controlled construction reduces variation in subjects
and scenes between paired images, helping focus the evaluation
on factual changes.
At test time, each image is presented independently, without
its paired counterpart or a verification target.
We report real- and fake-image accuracy, together with
\emph{pair accuracy}, which requires both images in a pair
to be classified correctly.
This metric exposes failures that can be hidden by a preference
for one label.

We also propose \textbf{PERSIST-Agent}, an iterative verification
framework built around a \emph{persistent verification state}.
The state records each candidate's entity and attribute,
visual observation, retrieved evidence, and verification status.
It separates what is observed in the image from what is supported
by evidence, while keeping both linked to the same verification target.
The agent uses this state to select candidates for further inspection
or retrieval and updates their statuses as new information becomes
available.
Unresolved or ambiguous candidates can be revisited rather than
being treated as settled.
This explicit record connects successive actions and provides
the basis for the final image-level judgment.
With backbone weights fixed, PERSIST-Agent performs
\emph{harness self-optimization}, iteratively refining its prompts
and execution rules through validation feedback.

Our experiments show strong label biases in several forensic models.
Among the models evaluated through direct judgment, the highest
pair accuracy is 32.20\%, leaving substantial room for improvement.
Retrieval improves pair accuracy, but the gains vary across models.
PERSIST-Agent improves over direct judgment by 10.40 and 9.41
percentage points on Qwen3-VL-8B and InternVL3.5-8B, respectively,
and outperforms the corresponding retrieval-augmented baselines.
Ablations further support the role of persistent state:
removing it reduces pair accuracy from 18.46\% to 11.86\%
on Qwen3-VL-8B.
These results highlight the importance of maintaining verification
state alongside access to external evidence.
They also show that reliable image-world consistency verification
remains an open challenge.

Our main contributions are as follows:
\begin{itemize}
    \item We introduce WorldFact-Bench to evaluate image-world
    consistency from a single image without a predefined verification
    target, using source-aligned pairs with traceable factual evidence.

    \item We propose PERSIST-Agent, which organizes iterative
    verification around a persistent state that links candidate facts,
    visual observations, evidence, and verification statuses.

    \item We evaluate forensic models and MLLM-based baselines,
    identify label biases and limited pair-wise discrimination,
    and show the benefits of state-guided verification through
    same-backbone comparisons and component ablations.
\end{itemize}

\section{Related Work}
\label{sec:related_work}

\paragraph{Image forensics and benchmarks.}
Image forensic methods use reconstruction errors, learned visual
representations, and semantic features to detect generated
or manipulated content~\citep{
Wang2023DIREFD,
Cazenavette2024FakeInversionLT,
Yan2024ASC,
Cheng2025COSPYCS}.
MLLM-based approaches further incorporate artifact explanations
and pattern-aware reasoning~\citep{Wen2025SpotTF,Tan2025VeritasGD}.
Their scope therefore extends beyond low-level texture cues.
Benchmarks such as GenImage and GIM evaluate cross-generator
detection and local manipulation detection,
respectively~\citep{Zhu2023GenImageAM,Chen2024GIMAM}.
WorldFact-Bench complements these settings by testing whether
visually plausible content conflicts with applicable facts or rules.
Its source-aligned pairs support evaluation of whether a detector
can accept an original image and reject its factually altered
counterpart.
Table~\ref{tab:benchmark-comparison} compares representative
forensic benchmarks.

\paragraph{Knowledge-grounded visual verification.}
Multimodal fact-checking uses external evidence to assess
image context and image--text claims~\citep{
Abdelnabi2021OpenDomainCM,
Wang2024MFCBenchBM,
Cao2025AVerImaTeCAD}.
Related benchmarks examine different aspects of visual factuality.
T2I-FactualBench evaluates whether text-to-image models depict
knowledge-intensive concepts specified by prompts~\citep{
Huang2024T2IFactualBenchBT}.
Pix2Fact studies fine-grained visual question answering
that requires external knowledge~\citep{Jiang2026Pix2FactWV}.
Recent work also extracts check-worthy claims from multimodal
social media posts~\citep{Teo2026MultimodalCE}.
WorldFact-Bench focuses on discovering and verifying factual
conflicts from an image alone.
It combines an unspecified verification target with controlled
factual manipulations and source-aligned pairs, allowing evaluation
of both target discovery and real--fake discrimination.

\paragraph{Verification agents and harness optimization.}
ReAct interleaves reasoning with external actions~\citep{
Yao2022ReActSR}, while FIRE combines evidence retrieval
and claim verification in an iterative process~\citep{
Xie2024FIREFW}.
These approaches establish a foundation for multi-step verification.
PERSIST-Agent organizes this process around an explicit,
candidate-level state table.
The table separates visual observations from evidence-supported
values, records unresolved candidates, and guides subsequent
inspection and retrieval.
Feedback-driven prompt optimization, as studied in
GEPA~\citep{Agrawal2025GEPARP}, also provides a basis for improving
agent instructions without updating model weights.
We use validation-driven harness self-optimization to refine
execution within a fixed state schema and action interface.
The central mechanism remains the persistent verification state
that connects candidate selection, evidence binding, and judgment.









\begin{table}[t]
\vspace{-3mm}
  \caption{Comparison with representative image forensic benchmarks.}
  \label{tab:benchmark-comparison}
  \centering

  \begingroup
  \scriptsize
  \setlength{\tabcolsep}{3pt}
  \renewcommand{\arraystretch}{0.94}
  \setlength{\aboverulesep}{1.5pt}
  \setlength{\belowrulesep}{1.5pt}

  \begin{tabular*}{\textwidth}{@{\extracolsep{\fill}}lccccccc@{}}
    \toprule
    \textbf{Benchmark}
    & \textbf{\shortstack{Full\\Gen.}}
    & \textbf{\shortstack{Local\\Edit}}
    & \textbf{\shortstack{Aligned\\Pair}}
    & \textbf{\shortstack{External\\Knowl.}}
    & \textbf{\shortstack{Factual\\Edit}}
    & \textbf{Reasoning}
    & \textbf{Loc.} \\
    \midrule

    GenImage~\cite{Zhu2023GenImageAM}
    & \checkmark & -- & -- & -- & -- & -- & -- \\

    GIM~\cite{Chen2024GIMAM}
    & -- & \checkmark & \checkmark
    & -- & -- & -- & \checkmark \\

    Community Forensics~\cite{Park2024CommunityFU}
    & \checkmark & -- & -- & -- & -- & -- & -- \\

    AI-GenBench~\cite{Pellegrini2025AIGenBenchAN}
    & \checkmark & -- & -- & -- & -- & -- & -- \\

    ForensicHub~\cite{Du2025ForensicHubAU}
    & \checkmark & \checkmark & $\triangle$
    & -- & -- & -- & \checkmark \\

    AEGIS~\cite{Zhang2026AEGISAH}
    & $\triangle$ & $\triangle$ & --
    & -- & -- & \checkmark & \checkmark \\

    \midrule
    \textbf{WorldFact-Bench}
    & -- & \checkmark & \checkmark
    & \checkmark & \checkmark & \checkmark & -- \\
    \bottomrule
  \end{tabular*}

  \par\vspace{1pt}
  \noindent
  \parbox{\textwidth}{\scriptsize
    \checkmark: explicit coverage;
    $\triangle$: partial coverage;
    --: not a primary setting.
  }
  \endgroup
\end{table}

\begin{figure}[t]
  \centering
  \includegraphics[width=\textwidth]
    {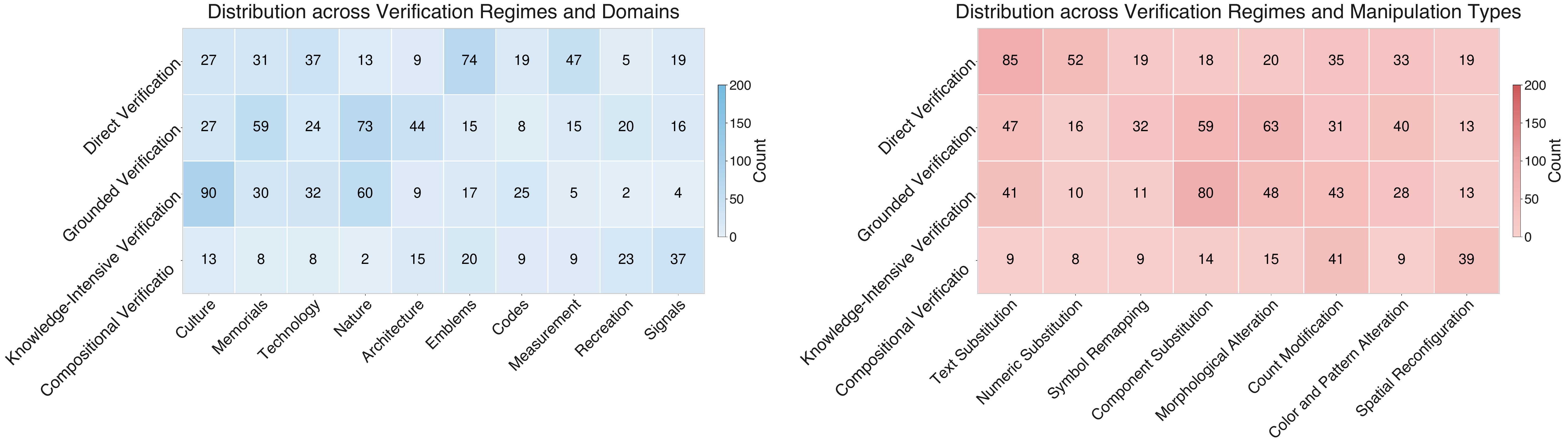}
    \vspace{-10mm}
  \caption{
    Distribution of the 1,000 test pairs across verification regimes
    and semantic domains (left) or manipulation types (right).
    Cell values indicate the number of real--fake pairs.
  }
  \vspace{-4mm}
  \label{fig:dataset-cross-coverage}
\end{figure}

\begin{figure*}[t]
    \centering
    \includegraphics[width=\textwidth]{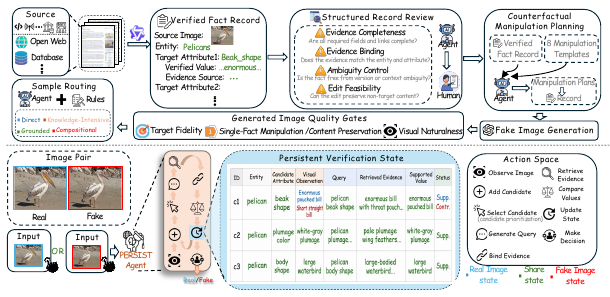}
    \caption{
Overview of WorldFact-Bench construction and PERSIST-Agent.
Top: verified fact records guide controlled image manipulation,
followed by quality checks and assignment to verification regimes.
Bottom: PERSIST-Agent verifies a single image through iterative
observation, retrieval, and comparison.
A persistent state table links candidate facts to visual observations,
evidence, and verification statuses, guiding subsequent actions
and retaining unresolved candidates.
The paired examples illustrate separate runs, not joint inputs.
}
    \vspace{-3mm}
    \label{fig:pipeline}
\end{figure*}

\section{WorldFact-Bench}
\label{sec:worldfact_bench}
\vspace{-2mm}
\subsection{Task Definition}
\label{sec:benchmark_overview}

WorldFact-Bench evaluates whether models can identify conflicts between image content and real-world facts or rules. Given a single image, a model predicts \textit{Real} or \textit{Fake}. It is not given a predefined verification target, reference evidence, or the paired image.
We construct pairs from the same source image. Starting with an original image whose target fact has been verified, we introduce one clear factual conflict through a local edit. The edit aims to preserve the subject, scene, and non-target content. The original and edited images are labeled \textit{Real} and \textit{Fake}, respectively, and evaluated independently. This design provides matched images that differ in the target fact, allowing us to test whether a model can both accept the original and detect the manipulation.

\subsection{Verification Regimes}
\label{sec:verification_regimes}

We organize samples into four regimes based on their verification requirements. \textbf{Direct Verification} compares a visual observation with a stable fact or rule when the target attribute and its context are explicit. \textbf{Grounded Verification} first requires identifying the relevant entity or target attribute from visual cues. \textbf{Knowledge-Intensive Verification} relies on detailed entity-specific facts, specialist knowledge, or archival records. \textbf{Compositional Verification} requires combining multiple visual observations and checking their relations or shared constraints. When several requirements apply, we assign one primary label in the following order: Compositional, Knowledge-Intensive, Grounded, and Direct. These labels describe verification requirements, not prescribed reasoning steps or tools.

\subsection{Dataset Construction and Quality Control}
\label{sec:dataset_construction}

\paragraph{Fact selection.}
We select visible, evidence-supported facts from images with traceable sources. For each candidate, we verify the relevant entity, attribute, and correct value. Images must provide sufficient visual cues for identification. We exclude candidates with missing context, severe occlusion, or unresolved ambiguity between variants.

\paragraph{Controlled manipulation.}
Each edit targets one fact and specifies both the region to modify and the content to preserve. The edited attribute or relation must contradict the relevant fact or rule, rather than represent another valid version, pose, or state. The edit must also preserve the visual cues needed to identify the entity and verify the fact.

\paragraph{Review and deduplication.}
Edited images undergo model-assisted checks and human review. We check whether the intended factual conflict is present, whether non-target content is preserved, and whether the edit introduces obvious unintended artifacts. We reject edits that fail to introduce the intended conflict, admit a valid alternative interpretation, or substantially alter non-target content. Before splitting the dataset, we remove duplicates through exact matching, perceptual similarity screening, and semantic review. Samples derived from the same source but targeting different facts may be retained; they are grouped together during splitting.

\subsection{Dataset Composition and Evaluation Protocol}
\label{sec:dataset_composition}
\label{sec:evaluation_protocol}

WorldFact-Bench contains 1,274 original--edited image pairs across four verification regimes, ten semantic domains, and eight manipulation types. Figure~\ref{fig:dataset-overview} presents category counts and representative examples. Figure~\ref{fig:dataset-cross-coverage} shows the distributions of verification regimes across semantic domains and manipulation types.

The dataset has no training split. We allocate 274 pairs to validation and 1,000 pairs to testing. Samples from the same source image or confirmed near-duplicate sources form a source group, which is kept within a single split. Subject to this constraint, the split aims to preserve the overall distributions along all three category dimensions while maintaining coverage of their pairwise combinations.

We report fake-image accuracy (F), real-image accuracy (R), and pair accuracy (Pair). F and R measure classification accuracy on edited and original images, respectively. Pair is our primary metric: a pair is counted as correct only if both images are classified correctly. A model that always predicts the same label therefore receives a Pair score of zero. All main results use the test split. Overall scores are computed over test samples, rather than as an unweighted average of category-level accuracies.

\begin{table}[t]
\vspace{-3mm}
\caption{
Main results on the WorldFact-Bench test set (\%).
F and R denote fake- and real-image accuracy, respectively;
Pair requires both images in a pair to be classified correctly.
The highest overall Pair accuracy is shown in bold.
}
\label{tab:main-results}
\centering
\scriptsize
\setlength{\tabcolsep}{3.1pt}
\renewcommand{\arraystretch}{0.92}
\resizebox{\linewidth}{!}{%
\begin{tabular}{@{}l cc cc cc cc ccc@{}}
\toprule
\multirow{2}{*}{\textbf{Method}} &
\multicolumn{2}{c}{\textbf{Direct}} &
\multicolumn{2}{c}{\textbf{Grounded}} &
\multicolumn{2}{c}{\textbf{K-Int.}} &
\multicolumn{2}{c}{\textbf{Compositional}} &
\multicolumn{3}{c}{\textbf{Overall}} \\
\cmidrule(lr){2-3}\cmidrule(lr){4-5}\cmidrule(lr){6-7}\cmidrule(lr){8-9}\cmidrule(lr){10-12}
& \textbf{F} & \textbf{R} & \textbf{F} & \textbf{R} & \textbf{F} & \textbf{R} & \textbf{F} & \textbf{R} & \textbf{F} & \textbf{R} & \textbf{Pair} \\
\midrule
\rowcolor{MethodSubgroupGray}
\multicolumn{12}{c}{\textbf{Specialized Vision Detectors}} \\
CO-SPY~\cite{Cheng2025COSPYCS}
& 22.69 & 81.76 & 0.00 & 100.00 & 0.00 & 89.47 & 0.00 & 94.74
& 6.38 & 91.23 & 1.83 \\
AIDE~\cite{Yan2024ASC}
& 0.00 & 58.43 & 0.00 & 100.00 & 4.00 & 100.00 & 10.53 & 89.47
& 2.61 & 86.80 & 2.61 \\
Effort~\cite{Yan2024EffortEO}
& 38.63 & 82.03 & 0.00 & 100.00 & 4.00 & 100.00 & 5.26 & 84.21
& 12.71 & 92.68 & 7.14 \\
PROBE~\cite{Cao2026WhereDF}
& 17.26 & 85.30 & 26.76 & 87.74 & 10.14 & 87.74 & 13.16 & 87.74
& 17.58 & 87.05 & 14.24 \\
\midrule
\rowcolor{MethodSubgroupGray}
\multicolumn{12}{c}{\textbf{MLLM-based Detectors}} \\
FakeVLM~\cite{Wen2025SpotTF}
& 0.00 & 100.00 & 0.00 & 100.00 & 0.00 & 100.00 & 0.00 & 100.00
& 0.00 & 100.00 & 0.00 \\
UniGenDet~\cite{Zhang2026UniGenDetAU}
& 100.00 & 0.00 & 100.00 & 0.00 & 100.00 & 0.00 & 100.00 & 0.00
& 100.00 & 0.00 & 0.00 \\
VERITAS~\cite{Tan2025VeritasGD}
& 0.00 & 100.00 & 0.00 & 97.78 & 22.00 & 78.95 & 5.26 & 94.74
& 6.79 & 92.81 & 0.57 \\
\midrule
\rowcolor{MethodGroupBlue}
\multicolumn{12}{c}{\textbf{MLLMs: Direct Judgment}} \\
Qwen3-VL-8B
& 22.38 & 77.48 & 0.00 & 100.00 & 28.00 & 86.84 & 31.58 & 73.68
& 18.51 & 86.28 & 8.06 \\
InternVL3.5-8B
& 82.24 & 21.96 & 15.15 & 91.11 & 90.00 & 5.26 & 73.68 & 15.79
& 62.94 & 37.31 & 8.86 \\
GLM-4.6V
& 0.00 & 100.00 & 0.00 & 100.00 & 8.00 & 92.11 & 15.79 & 89.47
& 4.47 & 96.32 & 3.64 \\
Kimi-K2.6
& 22.05 & 77.31 & 3.03 & 100.00 & 14.00 & 92.11 & 42.11 & 84.21
& 17.01 & 89.19 & 12.90 \\
MiniMax-M3
& 21.50 & 82.77 & 0.00 & 100.00 & 22.00 & 97.37 & 36.84 & 94.74
& 17.37 & 93.68 & 16.76 \\
Gemini-3.5-Flash
& 42.77 & 100.00 & 16.67 & 95.56 & 52.00 & 94.74 & 31.58 & 84.21
& 35.83 & 94.95 & \textbf{32.20} \\
\midrule
\rowcolor{MethodGroupBlue}
\multicolumn{12}{c}{\textbf{MLLMs: Retrieval-Augmented Judgment}} \\
InternVL3.5-8B + Retrieval
& 81.98 & 17.28 & 30.30 & 64.44 & 40.00 & 50.00 & 68.42 & 31.58
& 52.97 & 42.50 & 10.06 \\
Qwen3-VL-8B + Retrieval
& 38.05 & 42.76 & 6.06 & 97.78 & 28.00 & 78.95 & 31.58 & 94.74
& 24.74 & 76.72 & 14.82 \\
Kimi-K2.6 + Retrieval
& 18.67 & 100.00 & 6.06 & 100.00 & 52.00 & 100.00 & 42.11 & 89.47
& 27.38 & 98.48 & \underline{26.89} \\
\midrule
\rowcolor{MethodGroupBlue}
\multicolumn{12}{c}{\textbf{PERSIST-Agent}} \\
PERSIST-Agent (Qwen3-VL-4B)
& 78.46 & 22.30 & 21.21 & 75.56 & 70.00 & 47.37 & 52.63 & 84.21
& 55.19 & 54.12 & 17.16 \\
PERSIST-Agent (InternVL3.5-8B)
& 21.72 & 61.79 & 22.73 & 91.11 & 70.00 & 39.47 & 68.42 & 42.11
& 41.98 & 61.67 & 18.27 \\
PERSIST-Agent (Qwen3-VL-8B)
& 62.86 & 62.39 & 19.70 & 73.33 & 70.00 & 39.47 & 42.11 & 78.95
& 48.84 & 61.79 & 18.46 \\
\bottomrule
\end{tabular}%
}
\vspace{-2mm}
\end{table}


\section{PERSIST-Agent}
\label{sec:persist-agent}

Verifying image--world consistency requires more than retrieving
information about an image.
First, the verification target is not given: the model must identify
which visible details to check.
Second, retrieved information must match the entity, attribute,
and context under examination before it can support a judgment.
Finally, verification may remain incomplete after a single observation
or search.
The model must retain unresolved candidates and revise earlier
judgments as new evidence becomes available.
These requirements call for a shared working state that connects
candidate selection, evidence gathering, and judgment revision.

We propose \textbf{PERSIST-Agent}, an image verification framework
built around a persistent verification state.
As shown in Fig.~\ref{fig:pipeline}, a structured state table links
each candidate fact to its visual observation, retrieved evidence,
and current status.
The agent uses these records to select what to inspect or retrieve
next, then updates the relevant entries with the resulting information.
This allows verification to proceed across multiple candidates
without discarding unresolved questions.
Each run takes a single image as input, without access to its
paired counterpart, edit annotations, or benchmark reference evidence.

\subsection{Persistent Verification State}
\label{sec:persistent-state}

The state table organizes verification around candidate facts.
Each row represents an entity attribute or a relation to be checked.
For example, the pelican in Fig.~\ref{fig:pipeline} may yield
candidates concerning its bill shape, plumage color, and body shape.

Each record contains a candidate ID, entity, candidate attribute,
visual observation, query, retrieved evidence, evidence value,
and verification status.
The \emph{visual observation} describes what is visible in the image.
The \emph{evidence value} describes the corresponding attribute
according to an applicable source.
These fields are kept separate: the former is obtained through
image inspection, whereas the latter is extracted from evidence
matched to the entity and its context.
For bill shape, for example, the agent compares the observed
morphology with the description provided by a relevant source.

Each candidate has one of four statuses:
\textbf{Unverified}, when verification is incomplete;
\textbf{Supported}, when applicable evidence agrees with the observation;
\textbf{Contradicted}, when applicable evidence conflicts with it;
and \textbf{Ambiguous}, when the available information leaves
multiple valid interpretations.
Missing evidence alone does not establish a contradiction.
Evidence concerning a different entity, version, period, or rule
system cannot directly support a judgment.

The table serves as working state rather than only a record of
past actions.
Unverified candidates remain available for inspection, ambiguous
candidates can receive additional observations or evidence, and
previous judgments can be revised when relevant new information
becomes available.
This preserves verification progress across iterations.

\subsection{State-Guided Iterative Verification}
\label{sec:iterative-verification}

\paragraph{Candidate discovery.}
The agent first inspects the image for verifiable entities, text,
numbers, symbols, component properties, and spatial relations.
It adds these observations to the state table as candidate facts.
A candidate need not appear visually unusual: a plausible detail
may still conflict with an external fact.
When an observation is unclear, the agent records the uncertainty
and may inspect the relevant region again.

\paragraph{Candidate selection and evidence retrieval.}
At each iteration, the agent selects a candidate based on visual
clarity, verifiability, and its potential impact on the final decision.
If the observed value remains uncertain, the agent performs further
visual inspection.
Otherwise, it generates a query using the entity, target attribute,
and relevant context, then retrieves supporting material.
Previous queries and evidence remain available in the table for
reuse and to reduce redundant searches.

\paragraph{Evidence binding and state update.}
Before using a retrieved source, the agent checks whether it matches
the candidate's entity, attribute, and relevant context.
It then extracts the evidence value, compares it with the visual
observation, and updates the candidate's status.
For a relation or rule involving multiple observations, the agent
first gathers the required visual information and then checks
whether it satisfies the shared constraint.
If the comparison remains ambiguous, the candidate is retained
for further inspection, query refinement, or additional retrieval.

These actions form an iterative process rather than a fixed,
single-pass sequence.
Each action updates the relevant records while preserving other
candidates and their verification progress.
New evidence may revise an earlier judgment, and the agent may
switch between candidates as their information needs change.

\subsection{Image-Level Decision}
\label{sec:forensic-decision}

The final decision is based on the candidates and their bound evidence.
The agent predicts \textit{Fake} when a clearly observed candidate
conflicts with reliable, context-matched evidence.
Otherwise, it predicts \textit{Real} once all high-priority candidates
have been resolved or the reasoning budget is exhausted.
The output includes the image-level label, its rationale, and
the relevant evidence, while unresolved candidates remain recorded
in the state.
Here, \textit{Real} means that no factual conflict has been established
under the available observations, evidence, and budget;
it does not imply that every detail in the image has been verified.

\begin{figure}[t]
  \centering

  \begin{minipage}[t]{0.485\textwidth}
    \vspace{0pt}
    \centering

    \includegraphics[
      width=\linewidth
    ]{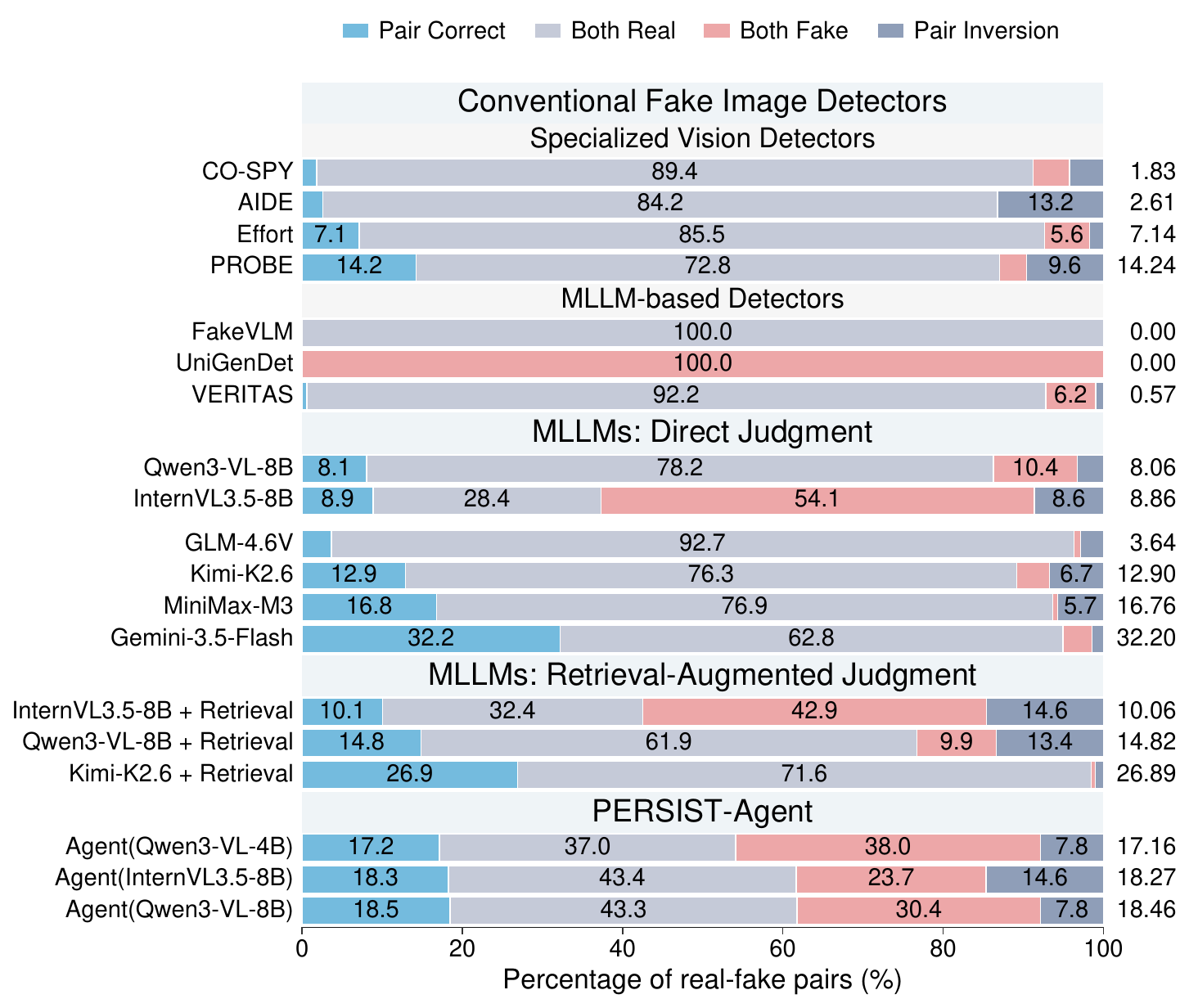}

    \captionof{figure}{
      Pair-wise prediction outcomes on the WorldFact-Bench test set.
      Pair Correct denotes two correct predictions;
      Both Real and Both Fake assign the same label to both images;
      Pair Inversion reverses both labels.
      Each bar shows the percentage of pairs in these four outcomes.
      Numbers on the right indicate Pair accuracy (\%).
    }
    \label{fig:pair-outcomes}
  \end{minipage}
  \hfill
  \begin{minipage}[t]{0.485\textwidth}
    \vspace{0pt}
    \centering

    \captionof{table}{
      Oracle analysis of PERSIST-Agent with Qwen3-VL-8B,
      measured by Pair accuracy (\%).
      Each partial oracle provides only the named gold information;
      All Gold provides entity, attribute, and evidence together.
    }
    \label{tab:oracle-diagnostics}

    \scriptsize
    \setlength{\tabcolsep}{2.5pt}
    \renewcommand{\arraystretch}{0.92}

    \resizebox{\linewidth}{!}{%
      \begin{tabular}{@{}lccccc@{}}
        \toprule
        \textbf{Setting}
        & \textbf{Direct}
        & \textbf{Grounded}
        & \textbf{K-Int.}
        & \textbf{Comp.}
        & \textbf{Overall} \\
        \midrule
        Base
        & 25.29 & 10.41 & 15.01 & 28.53 & 18.46 \\
        + Entity
        & 22.64 & 25.83 & 18.76 & 30.47 & 23.66 \\
        + Attribute
        & 25.91 & 38.54 & 26.18 & 40.72 & 31.92 \\
        + Evidence
        & 36.48 & 52.67 & 60.35 & 51.26 & 50.02 \\
        + All Gold
        & 69.24 & 74.18 & 80.63 & 76.47 & \textbf{74.89} \\
        \bottomrule
      \end{tabular}%
    }

    \vspace{3mm}

    \captionof{table}{
      Component ablation of PERSIST-Agent with Qwen3-VL-8B,
      measured by Pair accuracy (\%).
    }
    \label{tab:component-ablation}

    \scriptsize
    \setlength{\tabcolsep}{2.5pt}
    \renewcommand{\arraystretch}{0.92}

    \resizebox{\linewidth}{!}{%
      \begin{tabular}{@{}lccccc@{}}
        \toprule
        \textbf{Setting}
        & \textbf{Direct}
        & \textbf{Grounded}
        & \textbf{K-Int.}
        & \textbf{Comp.}
        & \textbf{Overall} \\
        \midrule
        Full
        & 25.29 & 10.41 & 15.01 & 28.53 & \textbf{18.46} \\
        w/o State
        & 18.92 & 5.88 & 8.64 & 16.73 & 11.86 \\
        w/o Prioritization
        & 19.47 & 7.82 & 10.96 & 21.44 & 13.92 \\
        w/o Binding
        & 18.35 & 6.43 & 7.28 & 23.61 & 12.49 \\
        w/o Ambiguity
        & 19.86 & 8.71 & 12.35 & 25.48 & \underline{15.26} \\
        \bottomrule
      \end{tabular}%
    }
  \end{minipage}
\vspace{-2mm}
\end{figure}

\section{Experiments}
\label{sec:experiments}

\subsection{Experimental Setup}
\label{sec:experimental-setup}

\paragraph{Compared methods.}
We evaluate five groups of methods: specialized vision detectors,
MLLM-based detectors, MLLMs with direct judgment, retrieval-augmented
MLLMs, and PERSIST-Agent across backbones.
Each method receives a single image without its paired counterpart,
a predefined claim, or the target attribute.
Real and fake images in each pair are evaluated independently.
Oracle experiments additionally provide the reference information
for each setting.

\paragraph{Data and metrics.}
We use 274 real--fake pairs for validation and 1,000 pairs for testing.
The test set contains 281 Direct, 301 Grounded,
274 Knowledge-Intensive, and 144 Compositional pairs.
Figure~\ref{fig:dataset-cross-coverage} shows their distribution
across semantic domains and manipulation types.
We report fake-image accuracy (F), real-image accuracy (R),
and pair accuracy (Pair).
A pair is correct only when both its real and fake images
are classified correctly.
Pair is our primary metric because it measures whether a method
can distinguish both images, rather than favoring one label.
All reported results are on the test set.
Oracle and ablation experiments use Qwen3-VL-8B.
All model results are averaged over three independent runs on the full test set to reduce stochastic variation.
\vspace{-2mm}
\paragraph{Harness optimization.}
We automatically optimize the agent harness using validation feedback
while keeping the backbone weights fixed.
The harness specifies the prompts and execution rules that guide
the agent's verification actions.
We retain the persistent state schema, candidate statuses,
and action interfaces, and optimize the instructions for candidate
selection, query generation, evidence binding, and ambiguity handling.
This process changes how the agent uses its available actions,
rather than the underlying model parameters.

Optimization proceeds iteratively.
At each iteration, the current harness is evaluated on validation
images, and its predictions and execution traces are collected.
An optimization model uses this feedback to propose revised
instructions or execution rules.
Candidate configurations are evaluated on the same validation samples
under the same per-image inference budget.
We select configurations primarily by validation Pair accuracy
and also inspect F and R to identify one-sided predictions.
The selected harness is fixed before test evaluation.
During testing, the agent updates only the verification state
of the current image; it does not revise the harness or use
test labels as feedback.

\subsection{Main Results}
\label{sec:main-results}

Table~\ref{tab:main-results} reports the main results.

\paragraph{Existing detectors struggle with factual manipulations.}
Specialized vision detectors achieve Pair scores between
1.83\% and 14.24\%, with PROBE performing best in this group.
Their real-image accuracy is consistently higher than their
fake-image accuracy, indicating a tendency to accept manipulated
images as real.
Among MLLM-based detectors, FakeVLM classifies every image as real,
whereas UniGenDet classifies every image as fake.
Both therefore obtain zero Pair accuracy despite achieving
100\% accuracy on one image class.
These results show that strong performance on one label
does not imply reliable discrimination between real and fake images.

\paragraph{Retrieval helps, but its gains vary across models.}
Adding retrieval improves Pair from 8.06\% to 14.82\% for Qwen3-VL-8B, from 8.86\% to 10.06\% for InternVL3.5-8B,
and from 12.90\% to 26.89\% for Kimi-K2.6.
These gains support the value of external evidence.
However, retrieval alone leaves substantial room for improvement.
The agent must still identify what to verify, retrieve relevant
information, and determine whether that information supports
or contradicts the visual observation.

\paragraph{PERSIST-Agent improves performance on the same backbones.}
With Qwen3-VL-8B, PERSIST-Agent achieves 18.46\% Pair,
improving over direct judgment and retrieval augmentation
by 10.40 and 3.64 percentage points, respectively.
With InternVL3.5-8B, it achieves 18.27\%, with corresponding
gains of 9.41 and 8.21 points.
The smaller Qwen3-VL-4B configuration achieves 17.16\%.
These results support the benefit of maintaining explicit
verification state across successive actions.
Nevertheless, Gemini-3.5-Flash achieves the highest Pair score
among evaluated methods at 32.20\%.
Image--world consistency remains challenging even for
the strongest models.

\subsection{Pair-wise Prediction Analysis}
\label{sec:pair-analysis}

Figure~\ref{fig:pair-outcomes} groups predictions into four outcomes:
\emph{Pair Correct}, where both images are classified correctly;
\emph{Both Real} and \emph{Both Fake}, where the same label is
assigned to both images; and \emph{Pair Inversion}, where both
predictions are incorrect.
This breakdown distinguishes verification
from one-sided label bias.
Specialized vision detectors predominantly produce
\emph{Both Real} predictions, while UniGenDet exhibits
the opposite pattern.
PERSIST-Agent reduces the dominant label bias of both
8B backbones.
For Qwen3-VL-8B, the proportion of \emph{Both Real} outcomes
decreases from 78.22\% under direct judgment to 43.33\%.
For InternVL3.5-8B, \emph{Both Fake} decreases from
54.08\% to 23.71\%.
Pair accuracy improves in both cases, although substantial
errors remain.
Thus, the benefit is not a more balanced prediction
distribution: PERSIST-Agent resolves more pairs.

\subsection{Diagnostic Analysis}
\label{sec:diagnostic-analysis}

\paragraph{Oracle analysis.}
Table~\ref{tab:oracle-diagnostics} examines performance when
reference entity, attribute, or evidence information is provided.
Compared with the base score of 18.46\%, the entity, attribute,
and evidence settings achieve overall Pair scores of
23.66\%, 31.92\%, and 50.02\%, respectively.
Evidence provides the largest gain among these partial-oracle
settings.
Its effect is particularly pronounced in Knowledge-Intensive
Verification, where Pair increases from 15.01\% to 60.35\%.
The benefits are not uniform across regimes.
Providing entity information improves Grounded Verification
from 10.41\% to 25.83\%, but reduces Direct Verification
from 25.29\% to 22.64\%.
Additional reference information therefore does not guarantee
an improvement for every verification regime.
Providing all gold information raises overall Pair to 74.89\%.
The remaining errors show that access to reference information
alone is insufficient for reliable final judgments.

\paragraph{Component ablation.}
Table~\ref{tab:component-ablation} evaluates the contribution
of each component.
Removing persistent state causes the largest overall decrease,
from 18.46\% to 11.86\%.
The drop is especially large in Compositional Verification,
where Pair falls from 28.53\% to 16.73\%.
This supports the role of persistent state in retaining
and combining observations across verification steps.
Removing evidence binding reduces overall Pair to 12.49\%,
with the largest regime-level decrease in Knowledge-Intensive
Verification, from 15.01\% to 7.28\%.
This highlights the importance of matching retrieved
evidence to the relevant entity and attribute.
Removing candidate prioritization or ambiguity handling
reduces overall Pair to 13.92\% or 15.26\%, respectively.
These results support the use of persistent state,
explicit evidence binding, and candidate management
within the verification process.

\section{Conclusion}
\label{sec:conclusion}

We introduced WorldFact-Bench to evaluate image-world consistency
beyond image-internal plausibility.
Its source-aligned real--fake pairs test whether models can
discover and verify factual conflicts from a single image
without a predefined target.
We also proposed PERSIST-Agent, which uses persistent verification
state to connect candidate discovery, visual observations,
evidence binding, and subsequent actions.
Experiments reveal substantial label biases and show that
retrieval alone yields uneven gains.
PERSIST-Agent improves over direct and retrieval-augmented
judgment on the evaluated 8B backbones, while ablations
support the role of persistent state.
Our evaluation focuses on controlled factual manipulations;
broader real-world coverage and reliable handling of unresolved
evidence remain important directions.
Together, the benchmark and framework support further study
of image verification grounded in both visual content
and real-world facts.

\section{AI Use Statement}

In this work, we used generative AI tools solely to polish the writing of the manuscript. We have reviewed all AI-assisted edits and take responsibility for the final content of this work, including all text, claims, and artifacts.

\bibliography{iclr2027_conference}
\bibliographystyle{iclr2027_conference}


\clearpage
\appendix
\raggedbottom

\section{Dataset Details}
\label{sec:app-dataset}

\subsection{Composition and split}
WorldFact-Bench contains 1,274 source-aligned real--fake pairs,
with 274 pairs for validation and 1,000 for testing.
These splits contain 548 and 2,000 images, respectively.
There is no training split.
Table~\ref{tab:app-coverage} gives the counts for all four verification
regimes, ten semantic domains, and eight manipulation types.
All categories in these three dimensions appear in both splits.
The three dimensions describe different properties of a sample:
the verification requirement, the image content, and the edit operation.
Their counts should therefore not be added across dimensions.

The original and edited images of a pair stay in the same split.
Images from the same source, including confirmed near-duplicate sources,
form a source group that is also kept within one split.
This constraint prevents the paired counterpart or a near-duplicate
source from appearing in validation when its test image is evaluated.
It does not require every entity, semantic category, or public rule
to be exclusive to one split.

\begin{table}[!htbp]
\caption{WorldFact-Bench composition by verification regime, semantic domain, and manipulation type. Counts refer to real--fake pairs. Each category dimension covers the same 1,274 pairs.}
\label{tab:app-coverage}
\centering\small
\setlength{\tabcolsep}{5pt}
\renewcommand{\arraystretch}{1.04}
\begin{tabular*}{\textwidth}{@{\extracolsep{\fill}}lrrr@{}}
\toprule
Category & Validation & Test & Total \\
\midrule
\multicolumn{4}{@{}l}{\textit{Verification regime}} \\
Direct & 77 & 281 & 358 \\
Grounded & 82 & 301 & 383 \\
Knowledge-Intensive & 76 & 274 & 350 \\
Compositional & 39 & 144 & 183 \\
\addlinespace[3pt]
\multicolumn{4}{@{}l}{\textit{Semantic domain}} \\
Culture & 42 & 157 & 199 \\
Memorials & 35 & 128 & 163 \\
Technology & 28 & 101 & 129 \\
Nature & 41 & 148 & 189 \\
Architecture & 21 & 77 & 98 \\
Emblems & 35 & 126 & 161 \\
Codes & 17 & 61 & 78 \\
Measurement & 20 & 76 & 96 \\
Recreation & 14 & 50 & 64 \\
Signals & 21 & 76 & 97 \\
\addlinespace[3pt]
\multicolumn{4}{@{}l}{\textit{Manipulation type}} \\
Text Substitution & 51 & 182 & 233 \\
Numeric Substitution & 23 & 86 & 109 \\
Symbol Remapping & 19 & 71 & 90 \\
Component Substitution & 47 & 171 & 218 \\
Morphological Alteration & 41 & 146 & 187 \\
Count Modification & 40 & 150 & 190 \\
Color and Pattern Alteration & 30 & 110 & 140 \\
Spatial Reconfiguration & 23 & 84 & 107 \\
\addlinespace[3pt]
\midrule
\textbf{All pairs} & \textbf{274} & \textbf{1,000} & \textbf{1,274} \\
\bottomrule
\end{tabular*}
\end{table}

\subsection{Assignment and quality checks}
A sample receives one primary verification regime.
When several requirements apply, the assignment order is
Compositional, Knowledge-Intensive, Grounded, and Direct.
This order resolves overlapping requirements; it is not an ordering
of difficulty or a prescribed sequence of model actions.
For example, a familiar object can require compositional verification
when a conflict can only be established by combining observations.
Conversely, a visually detailed image can admit a direct check when
the relevant value and rule are explicit.

An edit must introduce a factual conflict, not merely a visible change.
A different pose, a valid model variant, or another admissible state
is not sufficient to define a fake image.
The target must be visible and supported by applicable evidence,
and the edit must preserve the cues needed to identify the entity
and interpret its context.
Review checks target realization, preservation of non-target content,
and obvious unintended artifacts.
These checks aim to make factual verification informative; they do
not establish that every edited image is free of all forensic traces.

\subsection{Construction filtering and edit recipes}
\begin{figure}[!htbp]
  \centering
  \includegraphics[width=\textwidth]{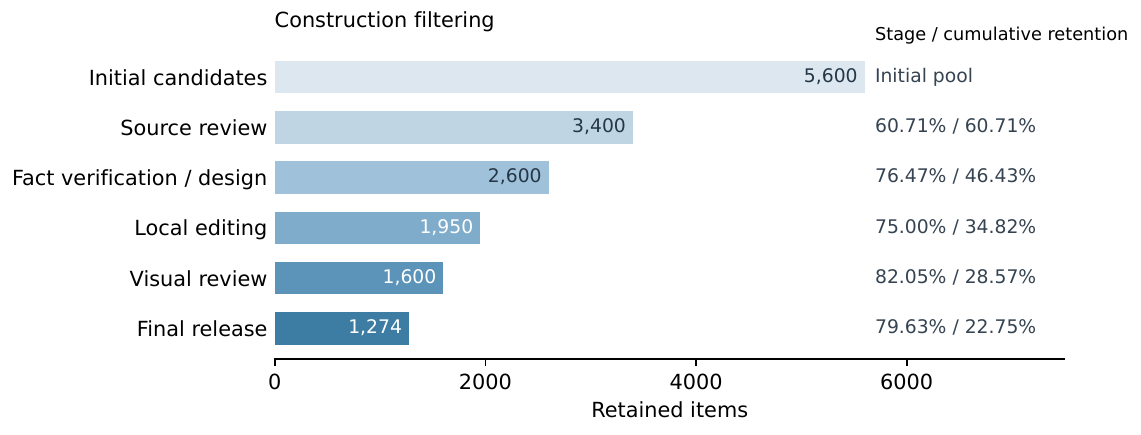}
  \caption{Construction filtering. Counts refer to source-image candidates before editing and real--fake pairs thereafter. The percentages give retention relative to the preceding stage and the initial pool, respectively. The final release contains 1,274 pairs.}
  \label{fig:app-filtering}
\end{figure}

Figure~\ref{fig:app-filtering} summarizes the construction stages.
The initial pool contains 5,600 candidates.
Source review retains 3,400, fact verification and counterfactual
design retain 2,600, and local editing yields 1,950 image pairs.
Visual review retains 1,600 pairs.
After the remaining review and deduplication, the released dataset
contains 1,274 pairs, or 22.75\% of the initial pool.
Rejection can occur when the evidence is insufficient, an edit
matches a valid alternative, the intended change is not realized,
or non-target content is substantially altered.

The construction library contains 221 atomic edit recipes.
These are reusable operations, not additional samples or target quotas.
A recipe specifies the target attribute, applicability conditions,
counterfactual change, preservation constraints, rejection conditions,
and acceptance checks.
Table~\ref{tab:app-recipes} summarizes the library.
The typical regime attached to a recipe is a construction tag;
the final regime is assigned to each sample using its actual
verification requirements.

\begin{table}[!htbp]
  \caption{Coverage of the 221 atomic edit recipes. The two groups classify the same recipe library by different criteria.}
  \label{tab:app-recipes}
  \centering\small
  \begin{tabular*}{\textwidth}{@{\extracolsep{\fill}}lrlr@{}}
    \toprule
    Manipulation type & Recipes & Typical verification regime & Recipes \\
    \midrule
    Text Substitution & 27 & Direct & 48 \\
    Numeric Substitution & 22 & Grounded & 41 \\
    Symbol Remapping & 24 & Knowledge-Intensive & 62 \\
    Component Substitution & 20 & Compositional & 70 \\
    Morphological Alteration & 17 & & \\
    Count Modification & 34 & & \\
    Color and Pattern Alteration & 18 & & \\
    Spatial Reconfiguration & 59 & & \\
    \midrule
    Total & 221 & Total & 221 \\
    \bottomrule
  \end{tabular*}
\end{table}

\section{Evaluation Protocol and Pair-wise Outcomes}
\label{sec:app-evaluation}

\subsection{Inputs and evaluation tracks}
Each standard evaluation receives one image at a time.
The paired image, reference evidence, and annotated verification target
are not supplied.
The pair association is used for scoring, not as model input.
Table~\ref{tab:app-tracks} summarizes the evaluated groups.
Oracle experiments use Qwen3-VL-8B and provide only the reference
information named in each setting; All Gold provides all three items.

\begin{table}[!htbp]
  \caption{Evaluation tracks used in the main paper. Standard settings receive a single image without a supplied verification target.}
  \label{tab:app-tracks}
  \centering\small
  \setlength{\tabcolsep}{4pt}
  \renewcommand{\arraystretch}{1.12}
  \begin{tabular}{@{}>{\raggedright\arraybackslash}p{0.29\textwidth}>{\raggedright\arraybackslash}p{0.67\textwidth}@{}}
    \toprule
    Track & Methods or backbones \\
    \midrule
    Specialized vision detectors & CO-SPY, AIDE, Effort, PROBE \\
    MLLM-based detectors & FakeVLM, UniGenDet, VERITAS \\
    Direct judgment & Qwen3-VL-8B, InternVL3.5-8B, GLM-4.6V, Kimi-K2.6, MiniMax-M3, Gemini-3.5-Flash \\
    Retrieval-augmented judgment & Qwen3-VL-8B, InternVL3.5-8B, Kimi-K2.6 \\
    PERSIST-Agent & Qwen3-VL-4B, InternVL3.5-8B, Qwen3-VL-8B \\
    \bottomrule
  \end{tabular}
\end{table}

\subsection{Inference configuration}
Retrieval-Augmented Judgment and PERSIST-Agent use the Serper
Google Search API under the shared limits in
Table~\ref{tab:app-inference}.
Direct judgment uses no external search.
We do not add retrieval or a persistent state to the public
forensic methods beyond their original implementations.
The shared limits bound available resources; they do not imply
that every method makes the same number of calls on each image.

\begin{table}[!htbp]
  \caption{Inference settings. Candidate and agent-step limits apply to PERSIST-Agent. Oracle variants retain the same limits and receive only their specified reference information.}
  \label{tab:app-inference}
  \centering\small
  \setlength{\tabcolsep}{4pt}
  \renewcommand{\arraystretch}{1.08}
  \begin{tabular*}{\textwidth}{@{\extracolsep{\fill}}ll@{}}
    \toprule
    Setting & Value \\
    \midrule
    Search service & Serper Google Search API \\
    Results per query & 10 \\
    Maximum search calls per image & 5 \\
    Maximum opened pages & 3 per query \\
    Maximum evidence spans & 12 \\
    Maximum span length & 256 tokens \\
    Maximum active candidates & 5 \\
    Maximum agent steps & 12 \\
    Decision temperature & 0 \\
    Image and reverse-image search & Disabled \\
    Search-result cache & Enabled, keyed by normalized query \\
    Output parsing & Fixed deterministic parser \\
    Invalid output & One retry; unresolved output is incorrect \\
    \bottomrule
  \end{tabular*}
\end{table}

Image and reverse-image search are disabled because retrieving
the source image could reveal the original value without discovering
the visual target.
Standard evaluation does not supply the verification regime label.
The parser maps generative outputs to Real or Fake.
An invalid output is retried once and counted as incorrect if
it remains invalid; it is not silently treated as a correct Real
prediction.

\subsection{Metrics and outcome decomposition}
For pair $i$, let $r_i$ indicate a correct prediction on the original
image and $f_i$ a correct prediction on the edited image.
For $N$ test pairs, the three accuracies are
\begin{equation}
  R=\frac{1}{N}\sum_{i=1}^{N}r_i,\qquad
  F=\frac{1}{N}\sum_{i=1}^{N}f_i,\qquad
  P=\frac{1}{N}\sum_{i=1}^{N}r_i f_i,
  \label{eq:app-metrics}
\end{equation}
where $P$ denotes Pair accuracy. Values in the tables are percentages.
Pair accuracy requires both predictions to be correct and is not
the mean or product of the two marginal accuracies.

For binary predictions, the four pair outcomes have probabilities
\begin{align}
  p_{\mathrm{correct}} &= P, &
  p_{\mathrm{both\ real}} &= R-P,\nonumber\\
  p_{\mathrm{both\ fake}} &= F-P, &
  p_{\mathrm{inversion}} &= 1-R-F+P.
  \label{eq:app-outcomes}
\end{align}
Here the accuracies are written as fractions in $[0,1]$.
The four probabilities sum to one.
Executions that remain invalid after the allowed retry are scored
as incorrect rather than omitted from the accuracy denominator.
A method that always predicts Real or always predicts Fake has
zero Pair accuracy, even though one of its marginal accuracies is one.
Table~\ref{tab:app-pair-outcomes} provides the complete outcome
decomposition corresponding to the main results.

\begin{table}[!htbp]
\caption{Complete pair-wise outcomes on the 1,000-pair test set (\%). The four outcomes are mutually exclusive and sum to 100\% for each method, up to rounding.}
\label{tab:app-pair-outcomes}
\centering\scriptsize
\setlength{\tabcolsep}{4pt}
\renewcommand{\arraystretch}{1.08}
\begin{tabular*}{\textwidth}{@{\extracolsep{\fill}}lrrrr@{}}
\toprule
Method & Pair Correct & Both Real & Both Fake & Pair Inversion \\
\midrule
\multicolumn{5}{@{}l}{\textit{Specialized Vision Detectors}} \\
CO-SPY & 1.83 & 89.40 & 4.55 & 4.22 \\
AIDE & 2.61 & 84.19 & 0.00 & 13.20 \\
Effort & 7.14 & 85.54 & 5.57 & 1.75 \\
PROBE & 14.24 & 72.81 & 3.34 & 9.61 \\
\addlinespace[3pt]
\multicolumn{5}{@{}l}{\textit{MLLM-based Detectors}} \\
FakeVLM & 0.00 & 100.00 & 0.00 & 0.00 \\
UniGenDet & 0.00 & 0.00 & 100.00 & 0.00 \\
VERITAS & 0.57 & 92.24 & 6.22 & 0.97 \\
\addlinespace[3pt]
\multicolumn{5}{@{}l}{\textit{MLLMs: Direct Judgment}} \\
Qwen3-VL-8B & 8.06 & 78.22 & 10.45 & 3.27 \\
InternVL3.5-8B & 8.86 & 28.45 & 54.08 & 8.61 \\
GLM-4.6V & 3.64 & 92.68 & 0.83 & 2.85 \\
Kimi-K2.6 & 12.90 & 76.29 & 4.11 & 6.70 \\
MiniMax-M3 & 16.76 & 76.92 & 0.61 & 5.71 \\
Gemini-3.5-Flash & 32.20 & 62.75 & 3.63 & 1.42 \\
\addlinespace[3pt]
\multicolumn{5}{@{}l}{\textit{MLLMs: Retrieval-Augmented Judgment}} \\
InternVL3.5-8B + Retrieval & 10.06 & 32.44 & 42.91 & 14.59 \\
Qwen3-VL-8B + Retrieval & 14.82 & 61.90 & 9.92 & 13.36 \\
Kimi-K2.6 + Retrieval & 26.89 & 71.59 & 0.49 & 1.03 \\
\addlinespace[3pt]
\multicolumn{5}{@{}l}{\textit{PERSIST-Agent}} \\
Qwen3-VL-4B & 17.16 & 36.96 & 38.03 & 7.85 \\
InternVL3.5-8B & 18.27 & 43.40 & 23.71 & 14.62 \\
Qwen3-VL-8B & 18.46 & 43.33 & 30.38 & 7.83 \\
\bottomrule
\end{tabular*}
\end{table}

Several detectors strongly favor one label.
CO-SPY assigns Real to both images for 89.40\% of pairs, and
AIDE does so for 84.19\%.
FakeVLM and UniGenDet show complete collapse toward Real and Fake,
respectively.
These outcomes distinguish one-sided prediction from reliable
discrimination between the two members of a pair.

\subsection{Comparison on the same backbones}
Table~\ref{tab:app-controlled} isolates the two 8B backbones evaluated
with direct judgment, retrieval augmentation, and PERSIST-Agent.
PERSIST-Agent improves Pair over direct judgment by 10.40 points
on Qwen3-VL-8B and 9.41 points on InternVL3.5-8B.
The gains over their retrieval-augmented variants are 3.64 and
8.21 points, respectively.
For Qwen3-VL-8B, the gap between F and R decreases from 67.77
to 12.95 points.
For InternVL3.5-8B, however, simple retrieval has a smaller gap
than PERSIST-Agent, despite its lower Pair accuracy.
Thus, balanced marginal accuracies alone do not establish successful
pair-level verification.

\begin{table}[!htbp]
\caption{Same-backbone comparison on the test set. F, R, and Pair are accuracies (\%). $\Delta$Pair is the gain over direct judgment in percentage points; Gap is $|F-R|$. A smaller gap alone does not imply higher pair accuracy.}
\label{tab:app-controlled}
\centering\scriptsize
\setlength{\tabcolsep}{3pt}
\renewcommand{\arraystretch}{1.1}
\begin{tabular*}{\textwidth}{@{\extracolsep{\fill}}llrrrrr@{}}
\toprule
Backbone & Setting & F & R & Pair & $\Delta$Pair & Gap \\
\midrule
Qwen3-VL-8B & Direct & 18.51 & 86.28 & 8.06 & -- & 67.77 \\
 & Retrieval & 24.74 & 76.72 & 14.82 & +6.76 & 51.98 \\
 & PERSIST-Agent & 48.84 & 61.79 & 18.46 & +10.40 & 12.95 \\
\addlinespace[2pt]
InternVL3.5-8B & Direct & 62.94 & 37.31 & 8.86 & -- & 25.63 \\
 & Retrieval & 52.97 & 42.50 & 10.06 & +1.20 & 10.47 \\
 & PERSIST-Agent & 41.98 & 61.67 & 18.27 & +9.41 & 19.69 \\
\addlinespace[2pt]
\bottomrule
\end{tabular*}
\end{table}

\section{Persistent State and Harness Optimization}
\label{sec:app-state}

\subsection{State semantics}
PERSIST-Agent separates an observation from the evidence used to
assess it.
A candidate record contains its ID, entity, candidate attribute,
visual observation, query, retrieved evidence, evidence value,
and verification status.
The visual observation records what the agent reads from the image.
The evidence value records what an applicable source supports.
A retrieval result should not silently replace an uncertain visual
observation with the value the agent expects to see.

Table~\ref{tab:app-status} specifies the meaning of each status.
Entity, attribute, and context matching are required before a source
can support either agreement or contradiction.
A source about a related product family, a different historical
period, or another rule system is not sufficient merely because
it is semantically relevant.
Likewise, failure to retrieve supporting evidence is not itself
evidence that the image is false.

\begin{table}[!htbp]
  \caption{Interpretation of candidate statuses in the persistent verification state. Statuses can be revised when new observations or evidence become available.}
  \label{tab:app-status}
  \centering\small
  \setlength{\tabcolsep}{4pt}
  \renewcommand{\arraystretch}{1.15}
  \begin{tabular}{@{}p{0.19\textwidth}p{0.77\textwidth}@{}}
    \toprule
    Status & Interpretation \\
    \midrule
    Unverified & The observations or evidence needed for a judgment are incomplete. \\
    Supported & Applicable evidence agrees with the visual observation. \\
    Contradicted & Applicable evidence conflicts with a clearly observed value or relation. \\
    Ambiguous & More than one valid interpretation remains, preventing a reliable comparison. \\
    \bottomrule
  \end{tabular}
\end{table}

Unverified and ambiguous candidates remain available for further work.
The agent can inspect the image again, refine a query, retrieve more
evidence, or switch to another candidate.
Updating one record does not discard the observations and evidence
already collected for other candidates.
For a relational check, the relevant observations must be considered
together before the rule is applied.
These properties make the table working state rather than only
a transcript of previous actions.

The final label follows the decision rule in the main paper.
A reliable factual contradiction supports Fake.
If no contradiction is established, the agent predicts Real after
resolving the high-priority candidates or exhausting its budget.
Unresolved candidates remain in the state.
Real therefore means that no conflict was established during that
run, not that every visible fact has been verified.

\subsection{What the harness optimization changes}
Harness self-optimization operates on the validation split.
The backbone weights, state schema, status definitions, and action
interfaces remain fixed.
The optimized instructions govern candidate selection, query
generation, evidence binding, and ambiguity handling.
The optimization model uses validation predictions and execution
traces to propose revisions, which are compared on the same
validation samples under the same per-image inference budget.
Validation Pair accuracy is the primary selection criterion;
F and R are also inspected for one-sided predictions.

This process is distinct from the agent's test-time state updates.
The selected harness is frozen before test evaluation.
During a test run, the agent updates only the verification state
for the current image; it does not optimize its prompts against
test labels or use the paired image as feedback.
This distinction separates validation-based optimization of the
verification procedure from evidence-driven reasoning on a test image.

\section{Additional Diagnostics}
\label{sec:app-diagnostics}

\subsection{Retrieval pipeline}
We evaluate retrieval traces from independently processed test images.
The Gold Target variant additionally receives the entity or rule
context and target attribute, but not gold evidence.
It uses the same search and reasoning limits as the standard agent.
We distinguish five stages of retrieval and evidence use:
\begin{itemize}
  \item \textbf{Query Target Accuracy (QTA):} at least one query
  targets the correct entity or rule context and attribute.
  \item \textbf{Gold Source Recall at 10 (GSR@10):} the recorded
  gold source appears among the top ten results of at least one query.
  \item \textbf{Equivalent Evidence Recall at 10 (EER@10):}
  a retrieved source supports the decisive fact, including a reliable
  alternative to the recorded gold source.
  \item \textbf{Opened Decisive Evidence Rate (ODER):} at least one
  opened page contains sufficient evidence for verification.
  \item \textbf{Evidence Binding Accuracy (EBA):} the selected evidence
  matches the entity, attribute, and relevant version, period, region,
  or rule context of the image.
\end{itemize}

Gold-source matching normalizes URLs, removes tracking parameters,
and accounts for redirects.
The other checks use the structured fact record and a shared rubric.
Aliases and equivalent paraphrases are accepted, but a context
mismatch is not.
An alternative source qualifies only if it supports the same decisive
value or rule.
If a required trace field is missing, the corresponding check is
counted as unsuccessful for that image.

\begin{table}[!htbp]
\caption{Retrieval diagnostics on the test set (\%). Gold Target supplies the entity or rule context and target attribute, but not reference evidence. All agent variants use Qwen3-VL-8B. Metric definitions are given in the text.}
\label{tab:app-retrieval}
\centering\small
\setlength{\tabcolsep}{4pt}
\renewcommand{\arraystretch}{1.07}
\begin{tabular*}{\textwidth}{@{\extracolsep{\fill}}lrrrrr@{}}
\toprule
Method / regime & QTA & GSR@10 & EER@10 & ODER & EBA \\
\midrule
\multicolumn{6}{@{}l}{\textit{Method comparison}} \\
Qwen3-VL-8B + Retrieval & 64.35 & 38.42 & 46.87 & 40.16 & 26.74 \\
PERSIST-Agent & 76.74 & 45.83 & 54.03 & 48.90 & 34.09 \\
PERSIST-Agent + Gold Target & 100.00 & 63.84 & 74.92 & 68.47 & 53.76 \\
\midrule
\multicolumn{6}{@{}l}{\textit{PERSIST-Agent by verification regime}} \\
Direct & 88.60 & 56.20 & 63.40 & 57.10 & 44.80 \\
Grounded & 54.10 & 31.20 & 37.10 & 32.40 & 20.30 \\
Knowledge-Intensive & 84.20 & 44.80 & 55.90 & 50.20 & 25.90 \\
Compositional & 86.70 & 58.10 & 67.60 & 64.90 & 57.60 \\
\midrule
Overall & 76.74 & 45.83 & 54.03 & 48.90 & 34.09 \\
\bottomrule
\end{tabular*}
\end{table}

Table~\ref{tab:app-retrieval} shows improvements over simple retrieval
at each measured stage.
PERSIST-Agent reaches 76.74\% QTA and 34.09\% EBA, compared with
64.35\% and 26.74\% for Qwen3-VL-8B with retrieval.
Its GSR@10, EER@10, and ODER are 45.83\%, 54.03\%, and 48.90\%.
Gold Target further raises EER@10 to 74.92\% and EBA to 53.76\%.
Thus, identifying the target helps but does not remove the need
to retrieve and bind applicable evidence.

Grounded Verification has the lowest QTA, 54.10\%.
Knowledge-Intensive Verification has substantially higher QTA,
84.20\%, but EBA remains at 25.90\%.
Compositional Verification has the highest EBA, 57.60\%, yet its
Pair accuracy is only 28.53\%.
These patterns distinguish target discovery from evidence acquisition
and from the subsequent visual comparison.

\subsection{Oracle interventions and component ablations}

\begin{figure}[!htbp]
  \centering
  \includegraphics[width=\textwidth]{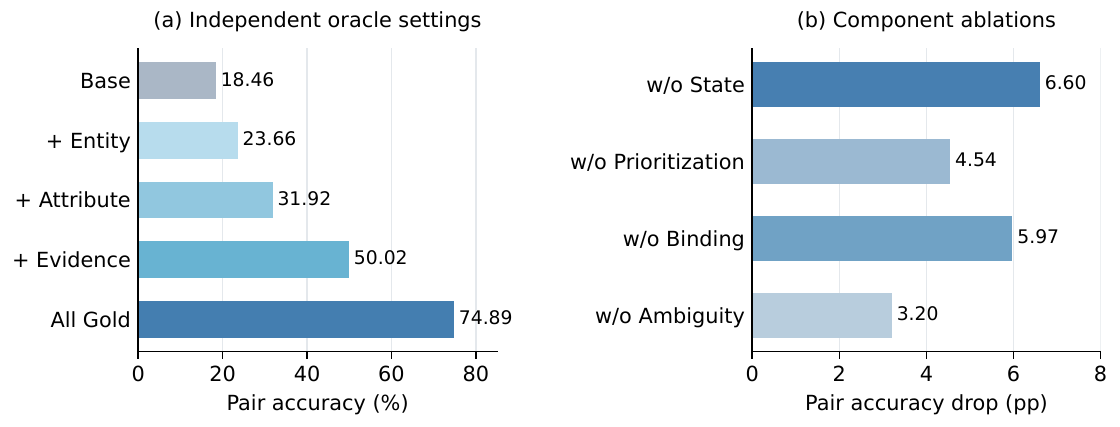}
  \caption{Oracle and ablation diagnostics for PERSIST-Agent with Qwen3-VL-8B. Left: Pair accuracy when gold information is supplied; each partial oracle provides only the named item, while All Gold provides entity, attribute, and evidence together. Right: Pair accuracy drops relative to the full agent (18.46\%) when individual components are removed.}
  \label{fig:app-diagnostics}
\end{figure}

Figure~\ref{fig:app-diagnostics} summarizes the oracle and ablation
results reported in the main paper.
The partial oracle settings are independent interventions, not a
sequence of cumulative additions.
Providing the entity, attribute, or evidence raises overall Pair
from 18.46\% to 23.66\%, 31.92\%, and 50.02\%, respectively.
These correspond to gains of 5.20, 13.46, and 31.56 percentage points.
All Gold supplies all three items and reaches 74.89\%.

The gains differ across verification regimes.
Gold Entity improves Grounded Verification from 10.41\% to 25.83\%,
but reduces Direct Verification from 25.29\% to 22.64\%.
Gold Evidence has its largest regime-level gain on
Knowledge-Intensive Verification, from 15.01\% to 60.35\%.
The remaining errors under All Gold show that access to reference
information does not, by itself, ensure a correct visual judgment.
The oracle scores do not identify the cause of every remaining error.

Removing persistent state produces the largest overall ablation drop,
6.60 points, followed by evidence binding at 5.97 points,
candidate prioritization at 4.54 points, and ambiguity handling
at 3.20 points.
The state ablation reduces Compositional Pair from 28.53\% to 16.73\%.
Removing evidence binding reduces Knowledge-Intensive Pair from
15.01\% to 7.28\%.
These differences support distinct roles for retaining observations
across steps and matching evidence to the right verification target.

\subsection{Primary failure causes}
We manually analyze 200 failed pairs from PERSIST-Agent with
Qwen3-VL-8B, with 50 pairs from each verification regime.
Each pair receives one primary failure label.
When several errors occur, the earliest identified error in the
verification chain determines the label, so a pair is not counted
under multiple causes.
This is an equal-size sample of failures from each regime,
not the full test-set distribution.

\begin{figure}[!htbp]
  \centering
  \includegraphics[width=\textwidth]{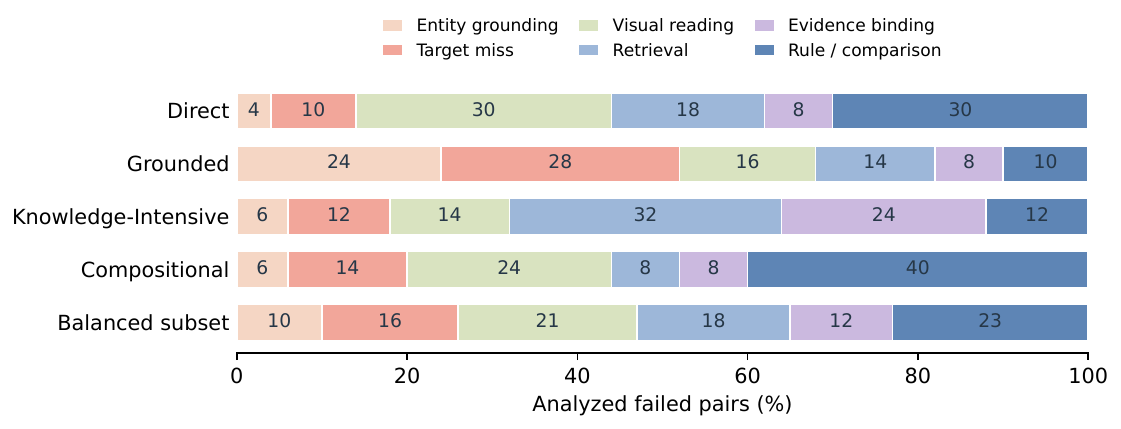}
  \caption{Primary failure causes in a manual audit of 200 failed pairs, with 50 pairs per verification regime. Each pair receives one label. The last bar summarizes this balanced failure subset, not the regime proportions of the full test set.}
  \label{fig:app-failures}
\end{figure}

In Direct Verification, visual reading and rule or comparison errors
each account for 30\% of the analyzed failures.
In Grounded Verification, entity grounding and target discovery
together account for 52\%.
Retrieval and evidence-binding errors account for 56\% in
Knowledge-Intensive Verification, while rule or comparison errors
account for 40\% in Compositional Verification.
Across the balanced subset, rule or comparison errors, visual reading
errors, and retrieval failures account for 23\%, 21\%, and 18\%,
respectively.
The analysis identifies recurring bottlenecks without treating these
subset percentages as population-wide failure rates.

\subsection{Agent behavior and resource use}
Table~\ref{tab:app-cost} summarizes the agent's behavior by regime.
Direct Verification has the shortest average process, with 1.30
search calls and 3.80 agent steps per image.
Knowledge-Intensive Verification uses the most search calls and
opened evidence pages, 3.70 and 7.20 on average.
Compositional Verification uses the most candidates and focused
inspections, 4.30 and 4.10, and has the largest mean step count, 7.80.
Overall, the agent uses 2.49 search calls and 6.05 steps per image.
These are operation counts, not measurements of latency or monetary cost.

\begin{table}[!htbp]
\caption{Agent behavior for PERSIST-Agent with Qwen3-VL-8B. The first five columns report per-image means. Ambig. is the percentage of images with an unresolved high-priority candidate at termination; Budget End is the percentage that reaches the reasoning limit.}
\label{tab:app-cost}
\centering\scriptsize
\setlength{\tabcolsep}{3pt}
\renewcommand{\arraystretch}{1.1}
\begin{tabular*}{\textwidth}{@{\extracolsep{\fill}}lrrrrrrr@{}}
\toprule
Regime & Candidates & \shortstack{Focused\\inspections} & \shortstack{Search\\calls} & \shortstack{Evidence\\pages} & \shortstack{Agent\\steps} & Ambig. & \shortstack{Budget\\End} \\
\midrule
Direct & 2.20 & 1.40 & 1.30 & 2.40 & 3.80 & 8.60 & 4.30 \\
Grounded & 3.50 & 2.80 & 2.60 & 4.70 & 6.10 & 18.90 & 12.80 \\
Knowledge-Intensive & 3.20 & 2.00 & 3.70 & 7.20 & 7.40 & 22.60 & 18.50 \\
Compositional & 4.30 & 4.10 & 2.30 & 4.30 & 7.80 & 20.10 & 16.20 \\
\midrule
Overall & 3.17 & 2.37 & 2.49 & 4.68 & 6.05 & 17.19 & 12.46 \\
\bottomrule
\end{tabular*}
\end{table}

The proportion reaching the reasoning limit is 12.46\% overall,
and 17.19\% retain an unresolved high-priority candidate at termination.
These observations complement final-label accuracy by describing
how much verification work remains incomplete.

\section{Qualitative Verification Examples}
\label{sec:app-qualitative}

The following diagrams summarize verification traces rather than
reproduce verbatim execution logs.
Paired images are shown together only for explanation; each is
processed independently.
The cases illustrate verification stages, not error frequencies.

\paragraph{A localized, checkable observation.}
In Figure~\ref{fig:app-keyboard}, the observed keyboard sequences are
\texttt{Q-W-E-R} and \texttt{Q-W-F-R}.
Matching this local observation to the QWERTY layout rule allows
the trace to accept the original and reject the altered image.

\begin{center}
\begin{minipage}{\textwidth}
  \centering
  \includegraphics[width=\textwidth]{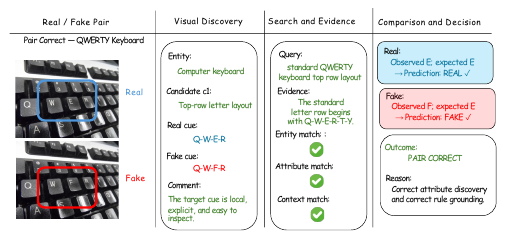}
  \captionof{figure}{A successful verification example on a QWERTY keyboard. The trace compares the observed top-row letters with the applicable layout rule and reaches a Pair Correct outcome. Blue and red boxes identify the corresponding regions in the original and edited images.}
  \label{fig:app-keyboard}
\end{minipage}
\end{center}

\paragraph{A relevant target with an overlooked visual change.}
Figure~\ref{fig:app-gauge} illustrates a Both Real outcome.
The original dial shows \texttt{3000}, while the edited image shows
\texttt{3500} at the same location.
The trace identifies the dial's numeric progression as relevant but
overlooks the altered mark and accepts both images.
This case shows that relevant evidence cannot compensate for an
incorrect local visual observation.

\begin{center}
\begin{minipage}{\textwidth}
  \centering
  \includegraphics[width=\textwidth]{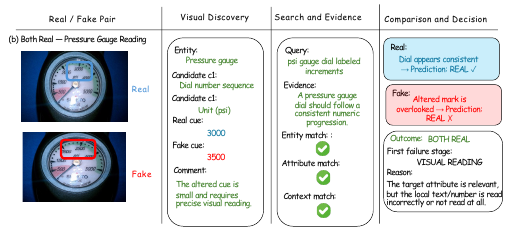}
  \captionof{figure}{A visual-reading failure on a pressure gauge. The target marking changes from 3000 in the original image to 3500 in the edited image. The trace overlooks this change and predicts Real for both, resulting in a Both Real outcome.}
  \label{fig:app-gauge}
\end{minipage}
\end{center}

\paragraph{Relevant evidence with the wrong scope.}
Figure~\ref{fig:app-camera} illustrates a Both Fake outcome.
Evidence about the Kodak Brownie family or nearby variants is applied
to an exact model designation, incorrectly rejecting the original image.
The case distinguishes retrieval relevance from valid evidence binding.

\begin{center}
\begin{minipage}{\textwidth}
  \centering
  \includegraphics[width=\textwidth]{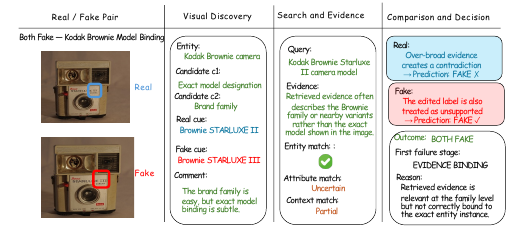}
  \captionof{figure}{An evidence-binding failure illustrated with a Kodak Brownie camera. Broad family-level evidence is applied to an exact model designation, resulting in a Both Fake outcome. Semantic relevance alone does not establish that a source applies to the depicted variant.}
  \label{fig:app-camera}
\end{minipage}
\end{center}

\paragraph{The correct rule applied to the wrong observation.}
In Figure~\ref{fig:app-domino}, the trace retrieves the relevant
matching rule but checks the wrong connection or miscompares its values.
It rejects the original and accepts the edited image.
This Pair Inversion illustrates the need to connect a rule to
the correct local observations.

\begin{center}
\begin{minipage}{\textwidth}
  \centering
  \includegraphics[width=\textwidth]{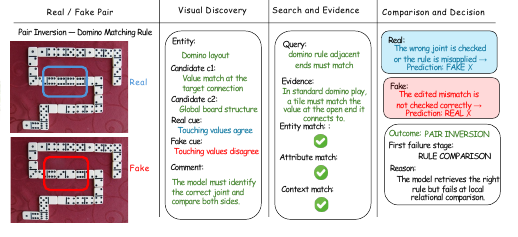}
  \captionof{figure}{A rule-comparison failure illustrated with a domino layout. The trace has the relevant matching rule but applies it to the wrong connection or miscompares the local values, leading to Pair Inversion.}
  \label{fig:app-domino}
\end{minipage}
\end{center}

\section{Scope and Limitations}
\label{sec:app-limitations}

Controlled, source-aligned edits do not cover all generated or
misleading imagery, especially cases that depend on missing event
context or an external textual claim.
Nor do the construction checks establish that every sample is free
of forensic traces.
A Real prediction can still leave candidates unresolved, so it is
not a guarantee of complete factual correctness.
Finally, the oracle and ablation results measure prediction changes,
not a complete causal attribution of errors; the manual failure
analysis describes only its balanced subset.

\end{document}